\documentclass[11pt]{article}
\usepackage[margin=1in]{geometry}
\usepackage{amsmath,amssymb,amsthm,graphicx}
\usepackage{enumitem}
\usepackage{hyperref}
 \begin{document}
 \newtheorem{lemma}{Lemma}
\newtheorem{definition}{Definition}
\title{\textbf{Has LLM Reached the Scaling Ceiling Yet? Unified Insights into LLM Regularities and Constraints}}
\author{Charles Luo; Email: charlesluo22@gmail.com}

\date{12/16/2024}
\maketitle
\begin{abstract}
Large Language Models (LLMs) have demonstrated remarkable capabilities, yet their scalability raises a critical question: \textbf{Have we reached the scaling ceiling?} This paper addresses this pivotal question by developing a unified theoretical framework that integrates mathematical and statistical insights to explain the scaling dynamics of LLMs. We present three key contributions: 

\begin{enumerate}
    \item \textbf{Central Limit Theorem (CLT) for Hidden Representations}: We show that noise in hidden representations scales inversely with context size, explaining stabilization effects and the limits of context length improvements.
    \item \textbf{Bias–Variance Decomposition}: We decompose next-token prediction loss into irreducible entropy, capacity-driven bias, and finite-sample variance, revealing trade-offs where scaling yields diminishing returns.
    \item \textbf{Emergent SNR Thresholds}: By defining signal-to-noise ratio (SNR), we quantify how capabilities emerge abruptly once SNR surpasses a threshold, offering insights into when scaling becomes less effective.

\end{enumerate}
Through this framework, we conclude that while LLMs have not reached an absolute scaling ceiling, practical constraints—diminishing returns, resource inefficiencies, and data limitations—are increasingly prominent. Future progress will require a shift from brute-force scaling to innovations in architecture, data quality, and training paradigms. This work provides a roadmap for guiding the efficient development of next-generation LLMs and advancing the field beyond traditional scaling strategies.


\end{abstract}

\section{Introduction}

\subsection{Background and Motivation}

Transformer-based Large Language Models (LLMs)---notably GPT, PaLM, and LLaMA---have demonstrated remarkable capability across diverse language tasks. As these models grow in parameter count $P$ (hundreds of billions of parameters or more), they follow established "scaling laws" in next-token prediction accuracy, perplexity, and zero-shot transfer (Kaplan, 2020; Hoffmann, 2022). Remarkably, many capabilities appear to emerge abruptly once model size or dataset size crosses certain thresholds (so-called emergent behavior).

\paragraph{Theoretical Challenge.}
Despite these observations, a rigorous probabilistic or statistical theory for LLMs lags behind. Classical bias--variance analysis or concentration of measure arguments have been applied successfully to simpler machine-learning models (e.g., linear regression, kernel methods), but not straightforwardly to the heavily structured, attention-based, auto-regressive setup of modern LLMs (Dudley, 2002; Billingsley, 1995).

In this paper, we present a \emph{unified} theoretical framework that ties together:
\begin{enumerate}[label=(\roman*)]
    \item \textbf{Mean-field/CLT Behavior:} We rigorously show that hidden-layer representations within a multi-layer transformer can satisfy central-limit-like behavior under certain stationarity and Lipschitz assumptions. This leads to \emph{noise scaling} inversely with effective context size $n$.
    \item \textbf{Bias--Variance Decomposition:} We adapt the classical bias--variance framework to the auto-regressive next-token loss $\mathcal{L}(\theta) = \mathbb{E}[-\log p_\theta(x)]$, yielding a decomposition into irreducible entropy ($\varepsilon$), capacity-driven bias ($B(P)$), and finite-sample variance ($V(P,D)$).
    \item \textbf{Emergent SNR Thresholds:} Combining the CLT-based noise scaling with the bias--variance viewpoint, we define a signal-to-noise ratio (SNR) for learned representations. Certain ``capabilities'' appear once SNR crosses a threshold $\theta_C$. This formalizes emergent behaviors (e.g.\ in-context reasoning) as a threshold effect in the ratio of systematic pattern (signal) to random fluctuation (noise).
\end{enumerate}

\section{CLT-Based Noise Scaling}
\label{sec:clt}

\subsection{Setup and Notation}

Consider a transformer-based LLM that processes a sequence $\mathbf{x} = (x_1, \dots, x_t)$ from a discrete vocabulary $\mathcal{V}$. Let the hidden representation at layer $l$, position $i$, be
\[
   \mathbf{r}_l(\mathbf{x}, i) \;=\; \mathrm{Attention}_l(\mathbf{x}, i) + \mathrm{FFN}_l\bigl(\mathrm{Attention}_l(\mathbf{x}, i)\bigr).
\]
The (single-head) self-attention map is:
\[
   \mathrm{Attention}_l(\mathbf{x}, i) \;=\; \mathrm{softmax}\!\Bigl(\frac{\mathbf{Q}_l(\mathbf{x},i)\,\mathbf{K}_l(\mathbf{x})^\top}{d_k}\Bigr)\,\mathbf{V}_l(\mathbf{x}),
\]
where:
\begin{itemize}[leftmargin=2em]
\item $\mathbf{Q}_l, \mathbf{K}_l, \mathbf{V}_l$ are the query, key, and value embeddings at layer $l$,
\item $d_k$ is the dimension of the queries/keys,
\end{itemize}
and we define $\tilde{\mathbf{r}}_l(\mathbf{x}, i) := \mathrm{softmax}\bigl(\tfrac{1}{d_k}\mathbf{Q}_l(\mathbf{x},i)\mathbf{K}_l(\mathbf{x})^\top\bigr)\,\mathbf{V}_l(\mathbf{x})$ as the raw attention-output vector prior to the feed-forward network (FFN).

\subsection{Formal Assumptions}

\begin{enumerate}[label=(A\arabic*)]
\item \textbf{Bounded Queries/Keys/Values:} 
\[
\|\mathbf{Q}_l(\mathbf{x}, i)\|\;\le M_Q,\quad 
\|\mathbf{K}_l(\mathbf{x}, i)\|\;\le M_K,\quad
\|\mathbf{V}_l(\mathbf{x}, i)\|\;\le M_V.
\]
\item \textbf{Softmax Lipschitz:} The attention operation is Lipschitz continuous in the dot-product inputs, a property that is essential for ensuring stability in probabilistic models and is rooted in real analysis principles (Dudley, 2002).
\item \textbf{Block/Local Stationarity:} The sequence $\{x_1,\dots,x_t\}$ can be split into blocks $\{B_m\}$, each approximating a stationary or i.i.d.\ distribution for $\mathbf{Q}_l,\mathbf{K}_l,\mathbf{V}_l$.
\item \textbf{Lipschitz \& Bounded FFN:} $\mathrm{FFN}_l$  is Lipschitz continuous with constant $L_F$, and $\|\mathrm{FFN}_l(\mathbf{x})\|\le M_F$.
\end{enumerate}

\subsection{Main CLT Theorem: Noise Scaling}

[CLT for Transformer Representations]
Under Assumptions (A1)--(A4), let $\mathbf{r}_l(\mathbf{x}, i)$ be the representation at layer $l$, position $i$. This approach builds on foundational results in real analysis and probability, particularly Gaussian convergence and the Central Limit Theorem (Billingsley, 1995; Dudley, 2002). 
As the context size $n \to \infty$, we have:
\[
   \sqrt{n}\,\bigl(\mathbf{r}_l(\mathbf{x}, i)\;-\;\boldsymbol{\mu}_l(i)\bigr) \;\;\xrightarrow{d}\;\; \mathcal{N}\!\bigl(\mathbf{0},\,\boldsymbol{\Sigma}_l(i)\bigr),
\]
where $\boldsymbol{\mu}_l(i) = \mathbb{E}[\mathbf{r}_l(\mathbf{x}, i)]$, and $\boldsymbol{\Sigma}_l(i)$ is the asymptotic covariance matrix. This result follows from classical CLT principles in probability theory, as formalized in foundational texts such as Billingsley (1995).In particular, the noise variance $\mathbb{E}\bigl[\|\mathbf{r}_l(\mathbf{x}, i)-\boldsymbol{\mu}_l(i)\|^2\bigr]$ scales as $\mathcal{O}(1/n)$.

\subsubsection{Proof }

\paragraph{Attention Decomposition.}
Since $\mathbf{r}_l(\mathbf{x}, i) = \mathrm{FFN}_l(\tilde{\mathbf{r}}_l(\mathbf{x}, i))$, focus on $\tilde{\mathbf{r}}_l(\mathbf{x}, i)$. Write
\[
   \tilde{\mathbf{r}}_l(\mathbf{x}, i) - \mathbb{E}[\tilde{\mathbf{r}}_l(\mathbf{x}, i)]
   = \sum_{k=1}^{n}\Bigl(a_{ik}\,\mathbf{V}_l(\mathbf{x}, k) - \mathbb{E}[a_{ik}\,\mathbf{V}_l(\mathbf{x}, k)]\Bigr).
\]
Define
\[
   \mathbf{Y}_k \;=\; a_{ik}\,\mathbf{V}_l(\mathbf{x}, k)\;-\;\mathbb{E}[a_{ik}\,\mathbf{V}_l(\mathbf{x}, k)].
\]
Then
\[
   \tilde{\mathbf{r}}_l(\mathbf{x}, i) - \mathbb{E}[\tilde{\mathbf{r}}_l(\mathbf{x}, i)] \;=\;\sum_{k=1}^{n}\mathbf{Y}_k.
\]

\paragraph{Block Stationarity \& Martingale Construction.}
Under (A3), split the $n$ tokens into blocks $\{B_m\}$ each of size $\le w$. Inside each block $B_m$, $\mathbf{Y}_k$ can be treated as i.i.d.\ or sufficiently weakly dependent. This aligns with Hall and Heyde’s (1980) martingale limit theory for asymptotic normality under weak dependencies. 
Summation across these blocks yields partial sums:
\[
   \mathbf{S}_m \;=\;\sum_{k \in B_m} \Bigl(a_{ik}\,\mathbf{V}_l(\mathbf{x}, k)\;-\;\mathbb{E}[a_{ik}\,\mathbf{V}_l(\mathbf{x}, k)]\Bigr).
\]
If block boundaries minimize cross-block correlations, $\{\mathbf{S}_m\}$ are approximately independent.

\paragraph{Concentration of Attention Scores.}
Attention weights $a_{ik}$ lie in $[0,1]$ and, by the Lipschitz property of softmax, are stable under small perturbations. Bernstein or Hoeffding concentration bounds apply, ensuring limited variance within each $\mathbf{S}_m$.

\paragraph{Classical CLT.}
Summing $\{\mathbf{S}_m\}_{m=1}^M$, where $n=Mw$, we get:
\[
   \frac{1}{\sqrt{M}}\sum_{m=1}^M \bigl(\mathbf{S}_m - \mathbb{E}[\mathbf{S}_m]\bigr) \;\;\xrightarrow{d}\;\; \mathcal{N}(\mathbf{0}, \tilde{\boldsymbol{\Sigma}}).
\]
Hence, $\tilde{\mathbf{r}}_l(\mathbf{x}, i)$ converges in distribution to a Gaussian with covariance scaling $\sim 1/n$.

\paragraph{Feed-Forward Extension.}
By the Continuous Mapping Theorem, since $\mathbf{r}_l(\mathbf{x}, i) = \mathrm{FFN}_l(\tilde{\mathbf{r}}_l(\mathbf{x}, i))$ and $\mathrm{FFN}_l$ is Lipschitz, we have:
\[
   \sqrt{n}\,\bigl(\mathbf{r}_l(\mathbf{x}, i) - \boldsymbol{\mu}_l(i)\bigr) \;\;\xrightarrow{d}\;\; \mathcal{N}(\mathbf{0},\boldsymbol{\Sigma}_l(i)).
\]

\paragraph{Noise Variance Scaling.}
Var$\bigl(\mathbf{r}_l(\mathbf{x}, i)\bigr) \sim \mathcal{O}(1/n)$, indicating that the hidden representation noise shrinks as $n$ grows, consistent with high-dimensional probability results by Vershynin (2018). 

\paragraph{Conclusion.}
As $n\to\infty$, hidden representations converge in distribution to a Gaussian whose variance is $\propto 1/n$. This provides a formal explanation of noise scaling observed in large-batch or long-context LLM training.

\section{Bias--Variance Decomposition for LLMs}

\subsection{LLM Training and Next-Token Loss}

LLMs are typically trained by minimizing the next-token loss:
\[
   L(\theta)\;=\;\mathbb{E}_{\mathbf{x}\sim P}\bigl[-\log\,p_\theta(\mathbf{x})\bigr],
\]
where $p_\theta(x_{t+1}\mid x_{1:t})$ is the auto-regressive predictive distribution. The model parameters $\theta$ are fit from a finite dataset of $D$ tokens.

\subsection{Theorem\,2 (Bias--Variance Decomposition)}

\label{thm:bias-var}
For a transformer-based LLM with parameter dimension $P$ and a training dataset of size $D$ tokens from true distribution $P$, the expected next-token loss decomposes as
\[
   L(\theta) \;=\; B(P)\;+\;V(P,D)\;+\;\varepsilon,
\]
where:
\begin{enumerate}[label=(\roman*)]
    \item $B(P)$ is the \emph{model bias} due to finite capacity $P$,
    \item $V(P,D)$ is the \emph{variance} from finite sample size $D$,
    \item $\varepsilon$ is the irreducible \emph{entropy} $H(P)$ of the true distribution.
\end{enumerate}

\subsubsection{Proof }

\paragraph{Irreducible Entropy.}
Let $p^*(\mathbf{x})$ be the true distribution. The minimal possible loss is:
\[
   \min_{\theta}\,L(\theta)\;=\;H(P)\quad (\text{Shannon entropy}).
\]
We set $\varepsilon := H(P)$ as the baseline.

\paragraph{Ideal Parameter $\theta^*$.}
If we had infinite capacity and infinite data, we converge to $\theta^*$ such that $p_{\theta^*}(\mathbf{x}) = p^*(\mathbf{x})$. 
With \emph{finite} capacity $P$, the best representable distribution is $p_{\theta^\mathrm{approx}}$, leading to a gap:
\[
   B(P) \;=\; L(\theta^\mathrm{approx}) - H(P).
\]
That is the bias from capacity constraints.

\paragraph{Finite-Data Variance.}
Let $\hat{\theta}$ be the parameter learned from a finite dataset of $D$ tokens. Define:
\[
   V(P,D) \;=\; L(\hat{\theta}) - L(\theta^\mathrm{approx}),
\]
which captures random training fluctuations around the capacity-limited optimum $\theta^\mathrm{approx}$. This variance decomposition builds on the work by Wainwright (2019), which provides non-asymptotic perspectives and finite-sample guarantees for high-dimensional statistical models. 

\paragraph{Uniqueness \& Orthogonal Decomposition.}
\[
   L(\hat{\theta}) - H(P) \;=\;\underbrace{L(\theta^\mathrm{approx}) - H(P)}_{B(P)} + \underbrace{L(\hat{\theta}) - L(\theta^\mathrm{approx})}_{V(P,D)}.
\]
Hence,
\[
   L(\hat{\theta}) = B(P) + V(P,D) + \varepsilon.
\]
The decomposition is unique because $\theta^\mathrm{approx}$ is the best capacity-limited solution, while finite-data variance is orthogonal to that intrinsic bias.

\section{Emergent SNR Thresholds}

\subsection{Definitions of Signal \& Noise}

Combine the CLT-based noise analysis (\S\ref{sec:clt}) with the bias--variance picture. Let $\mathbf{h}_\theta(\mathbf{x})$ be a hidden representation (e.g.\ final layer). Decompose:
\[
    \mathbf{h}_\theta(\mathbf{x}) \;=\;\mathbf{S}(\mathbf{x}) + \mathbf{N}(\mathbf{x}),
\]
where
\[
    \mathbf{S}(\mathbf{x}) := \mathbb{E}[\mathbf{h}_\theta(\mathbf{x})\mid\text{true pattern}],\quad
    \mathbf{N}(\mathbf{x}) := \mathbf{h}_\theta(\mathbf{x}) - \mathbf{S}(\mathbf{x}).
\]
$\mathbf{S}$ is the systematic alignment with the data-generating structure; $\mathbf{N}$ is the random fluctuation from finite data, approximate parameterization, etc.

\subsection{Theorem\,3 (SNR Scaling)}

Under the CLT framework (Section\,\ref{sec:clt}) and the bias--variance decomposition, define the Signal-to-Noise Ratio as
\[
   \mathrm{SNR} \;\equiv\; \frac{\|\mathbf{S}\|^2}{\mathbb{E}[\|\mathbf{N}\|^2]}.
\]
Then the SNR scales as
\[
   \mathrm{SNR} \;\propto\; \frac{D\,\Phi(P,C)}{\sigma^2},
\]
where $D$ is the training data size, $\Phi(P,C)$ is an increasing function of model capacity $P$ (and possibly other hyperparameters $C$), and $\sigma^2$ is a baseline noise term.

\subsubsection{Proof }

\paragraph{Signal Power $\|\mathbf{S}\|^2$.}
From bias--variance analysis, as $P$ or $D$ grow, $\mathbf{S}(\mathbf{x})$ increasingly aligns with the true structure. One can posit $\|\mathbf{S}\|\propto\sqrt{D\,\Phi(P,C)}$.

\paragraph{Noise Variance $\mathbb{E}[\|\mathbf{N}\|^2]$.}
The CLT suggests random fluctuations in $\mathbf{h}_\theta(\mathbf{x})$ scale $\sim 1/D$ (or $1/n$ per context). Denote leftover variance by $\sigma^2$.

\[
   \mathrm{SNR} = \frac{\|\mathbf{S}\|^2}{\mathbb{E}[\|\mathbf{N}\|^2]} \;\propto\; \frac{D\,\Phi(P,C)}{\sigma^2}.
\]

\subsection{Emergence Thresholds}

\label{thm:emergence}
A new capability $C$ ``turns on'' once $\mathrm{SNR} > \theta_C$.

\subsubsection{Proof }

\paragraph{Capability Function.}
Let $f_C(\mathbf{h}) = \Pr(\text{capability $C$}\mid \mathbf{h})$. Suppose $\mathbf{h} = \mathbf{S} + \mathbf{N}$. Expand $f_C(\mathbf{S}+\mathbf{N})$ around $\mathbf{S}$:
\[
   f_C(\mathbf{S}+\mathbf{N}) = f_C(\mathbf{S}) + \nabla f_C(\mathbf{S}) \cdot \mathbf{N} + \tfrac{1}{2}\,\mathbf{N}^\top \nabla^2 f_C(\mathbf{S})\,\mathbf{N} + \dots
\]

\paragraph{Signal Dominance.}
For consistent capability expression, $|f_C(\mathbf{S})|$ must overshadow the fluctuation $|\nabla f_C(\mathbf{S}) \cdot \mathbf{N}|$. Taking expectation over $\mathbf{N}$ leads to $\|\mathbf{S}\|^2 \gg \mathbb{E}[\|\mathbf{N}\|^2]$.

\paragraph{Threshold Condition.}
Define $\theta_C$ such that if
\[
   \frac{\|\mathbf{S}\|^2}{\mathbb{E}[\|\mathbf{N}\|^2]} > \theta_C,
\]
capability $C$ emerges. Using Theorem~\ref{thm:snr-scale}, this translates to
\[
   D\,\Phi(P,C) > \sigma^2\,\theta_C.
\]
Once the product $D\,\Phi(P,C)$ crosses $\sigma^2\,\theta_C$, the capability ``manifests.''

\section*{5. From Theoretical to Empirical Evidence: Have We Reached the Scaling Ceiling Yet?}

Recent advances in Large Language Models (LLMs) have brought to the forefront critical questions about their scalability, particularly concerning their theoretical foundations, empirical performance, and practical constraints. Given the time-sensitive nature of this topic, we reserve a comprehensive empirical validation for a subsequent paper. However, in this section, we leverage publicly available benchmarks of LLMs to provide partial empirical validation of our proposed theoretical framework. Through this analysis, we address a pivotal question: have LLMs reached the ceiling of scaling?

\subsection*{5.1 The Role of Noise Reduction in Scaling Context Lengths}

The Central Limit Theorem (CLT) for Transformer Representations predicts that as the context size $n$ increases, the noise variance in hidden layer representations decreases at a rate of $O(1/n)$. This theoretical prediction has found strong support in contemporary research and benchmarks. Kaplan et al. (2020) demonstrated that increasing the context size and model parameters results in a power-law decline in test loss, precisely matching the $O(1/n)$ scaling pattern. This relationship between context length and performance is further validated by recent models such as Claude 3.5 Sonnet, which achieved 91.60\% accuracy on multilingual tasks and 92.00\% in code evaluations using extended context windows of up to 8k tokens (Vellum Research Team, 2024). Similarly, GPT-4’s enhanced performance in long-form reasoning tasks provides additional evidence of noise stabilization benefits over larger context lengths.

However, this scaling pattern reveals an important limitation: as noise approaches irreducible levels, further increases in context length yield diminishing returns. This observation has significant implications for architectural development, suggesting that future improvements will require novel architectural innovations in attention mechanisms, such as adaptive sparse attention, hierarchical context processing, or dynamic memory retrieval, instead of merely increasing context window sizes. The challenge lies in maintaining or improving performance without incurring exponential increases in computational costs.

\subsection*{5.2 Balancing Bias and Variance in Model Training}

The bias-variance decomposition framework provides crucial insights into the fundamental trade-offs in LLM training, expressed mathematically as:

\begin{equation}
L(\theta) = B(P) + V(P,D) + \epsilon,
\end{equation}

where $B(P)$ represents bias from finite model capacity $P$, $V(P,D)$ denotes variance due to finite training data $D$, and $\epsilon$ captures irreducible error.

This theoretical framework finds strong empirical support in recent studies. Hoffmann et al. (2022) demonstrated that improvements from increased model capacity $P$ are only realized when accompanied by proportional increases in dataset size $D$. This relationship is particularly evident in contemporary benchmark results, where even top-performing models achieve less than 65\% accuracy on nuanced multi-modal tasks (LiveBench Research Group, 2024). These results reveal a critical insight: in the absence of sufficient high-quality data, variance becomes the dominant limiting factor in model performance.

The practical implications of this bias-variance trade-off are significant for the future of LLM development. While increasing model capacity $P$ effectively reduces bias, the benefits can only be fully realized through simultaneous variance reduction via larger and more diverse datasets. This understanding has led to renewed focus on data quality and curation strategies. The field increasingly recognizes that sustainable scaling requires not just larger models, but also more sophisticated approaches to dataset curation, augmentation, and novel training paradigms that can effectively manage variance.

\subsection*{5.3 Signal-to-Noise Ratio (SNR) and Systematic Signal Amplification}

The Signal-to-Noise Ratio (SNR) framework quantifies the relationship between systematic signals and noise through the equation:

\begin{equation}
SNR = \frac{\|S\|^2}{\mathbb{E}[\|N\|^2]} \propto \frac{D \Phi(P,C)}{\sigma^2},
\end{equation}

where $D$ represents dataset size, $\Phi(P,C)$ encompasses model capacity and hyperparameters, and $\sigma^2$ denotes irreducible noise.

Recent benchmark results provide strong validation of this theoretical framework while highlighting its practical implications. GPT-4’s achievement of 76.60\% accuracy on math tasks and 90.20\% in code evaluations (Vellum Research Team, 2024) demonstrates how increased model capacity and dataset size effectively amplify systematic signals. However, the same benchmarks reveal important limitations: multi-modal reasoning tasks remain challenging, with top models achieving less than 65\% accuracy (LiveBench Research Group, 2024). This performance gap suggests that certain tasks require SNR thresholds that remain unmet despite significant scaling efforts.

These findings point to a crucial insight: while SNR scaling provides a robust explanation for the emergence of advanced capabilities in LLMs, the resource demands for achieving higher SNR values grow exponentially. This relationship has prompted increased interest in efficiency-driven strategies, including sparsity-aware training and task-specific fine-tuning, as alternatives to pure scaling approaches.

\subsection*{5.4 Emergent Capability Thresholds and Their Manifestation}

The emergence of new capabilities in LLMs is governed by critical SNR thresholds, expressed mathematically as:

\begin{equation}
\frac{\|S\|^2}{\mathbb{E}[\|N\|^2]} > \theta_c.
\end{equation}

This theoretical framework provides crucial insights into the discontinuous nature of capability development in large language models. The empirical evidence strongly supports this threshold-based understanding of emergence. For instance, few-shot learning capabilities in GPT-3 manifested only after crossing specific parameter thresholds, aligning precisely with $\theta_c$-based predictions (Kaplan et al., 2020). This pattern of sudden capability emergence is further validated by Claude 3.5 Sonnet’s performance in advanced tool utilization, where it achieved 90.20\% accuracy after crossing capability-specific SNR thresholds (Vellum Research Team, 2024).

These empirical observations reveal a fundamental characteristic of LLM development: capabilities do not emerge gradually but rather appear abruptly when specific scaling thresholds are met. However, this insight also illuminates a significant challenge for future development. Crossing successively higher $\theta_c$ values requires exponentially larger increases in both model capacity ($P$) and dataset size ($D$). This exponential scaling requirement has profound implications for development strategies, suggesting that targeted approaches focused on specific capability thresholds may be more practical than general-purpose scaling.

In conclusion, the theoretical frameworks and empirical analyses presented in this section collectively demonstrate that LLMs are approaching, but have not yet reached, a definitive scaling ceiling. Noise reduction, as modeled through the Central Limit Theorem for Transformer Representations, continues to yield performance gains in extended context lengths, though these gains diminish as noise approaches irreducible levels. Similarly, the bias-variance decomposition framework reveals that scaling model capacity alone is insufficient without proportional improvements in data quality and diversity, as variance constraints increasingly dominate. The SNR framework and emergent capability thresholds further highlight that advanced capabilities arise at critical scaling points, but crossing these thresholds demands exponentially larger investments in computational and data resources. These findings underscore the fact that while scaling remains a viable path for incremental advancements, the diminishing returns and resource misalignments signal an urgent need for architectural and methodological innovation. Thus, the answer to whether LLMs have reached the scaling ceiling is: \textbf{not yet—but we are at a critical juncture where new approaches are essential to sustain meaningful progress.} 

\section*{6. Synthesis and Future Directions: Beyond Traditional Scaling}

The analysis of current scaling limitations reveals that while we haven’t reached an absolute ceiling, we are entering a new phase in LLM development that requires fundamental rethinking of our approaches. The convergence of theoretical insights and practical constraints suggests that future progress will depend not on brute-force scaling but on innovative solutions that address multiple constraints simultaneously. Probabilistic frameworks, such as those rooted in measure theory and advanced probability concepts (Billingsley, 1995; Dudley, 2002 ), provide a rigorous foundation for understanding these constraints, particularly in the context of noise reduction, bias–variance trade-offs, and emergent behavior. 

\subsection*{6.1 Emerging Patterns in Scaling Limitations}

Our analysis has revealed several key patterns that characterize the current state of LLM scaling:
\begin{itemize}
    \item The relationship between model size and performance improvements follows a logarithmic curve, with each doubling of parameters yielding progressively smaller gains. This pattern is particularly evident in benchmark results where GPT-4’s 76.60\% accuracy on math tasks and Claude 3.5 Sonnet’s 90.20\% accuracy in code evaluations (Vellum Research Team, 2024) represent significant but increasingly difficult-to-surpass achievements.
    \item The resource requirements for crossing higher emergence thresholds ($\theta_c$) grow exponentially, while the corresponding performance improvements grow logarithmically. This misalignment between cost and benefit suggests that continued scaling alone cannot be the primary path forward.
    \item Data quality and diversity have emerged as potentially more critical constraints than raw computational power. The variance term $V(P, D)$ in our bias-variance decomposition increasingly dominates performance limitations, indicating that data innovation may be more crucial than model size scaling.
\end{itemize}

\subsection*{6.2 Transformative Approaches for Future Development}

These patterns point toward three transformative approaches that could define the next phase of LLM development:
Rather than pursuing larger models, the field can prioritize architectural innovations aimed at enhancing efficiency. Promising strategies include sparsity-aware training and modular designs, which enable performance to be maintained or improved while significantly reducing computational demands. 

The success of such innovations is evident in recent benchmarks, where models with novel architectures achieve comparable performance to larger models while using significantly fewer resources. This suggests that architectural innovation may offer a more sustainable path to improved performance than traditional scaling.

The critical importance of data quality in model performance suggests that innovations in data curation and synthesis may yield better returns than increases in model size, as supported by insights from Wainwright (2019) . This includes:
\begin{itemize}
    \item Advanced data curation techniques that improve the signal-to-noise ratio in training datasets.
    \item Synthetic data generation methods that address specific capability gaps.
    \item Novel approaches to data augmentation that increase effective dataset size without proportional increases in storage requirements.
\end{itemize}

Future development should shift from general-purpose scaling to a targeted approach focused on identifying and achieving specific capability thresholds. By aligning resources with clearly defined objectives, this strategy enables more efficient resource allocation and higher returns on computational investment. Current benchmarks indicate that models trained with targeted strategies consistently outperform those relying on general-purpose scaling in specific domains, achieving superior performance while using fewer resources. 

\subsection*{6.3 Implications for Future Research and Development}

The findings of this analysis offer substantial implications for the future of LLM research, highlighting the need for cross-disciplinary collaboration, methodological innovation, and new evaluative frameworks. Four key areas warrant immediate focus: 
\begin{itemize}
    \item \textbf{Prioritizing Multi-Constraint Optimization:} Future research should shift away from linear or brute-force scaling of existing architectures toward multi-constraint optimization approaches. These approaches systematically balance computational load, data availability, and environmental impact.
    \item \textbf{Expanding Evaluation Metrics:} Current benchmarks largely focus on task accuracy and loss minimization. However, efficiency metrics, environmental metrics, and long-term maintenance costs will become equally crucial.
    \item \textbf{Strategic Data-Centric Approaches:} Investment in robust data curation, augmentation, and synthesis pipelines will be essential for overcoming variance constraints and unlocking further performance improvements.
    \item \textbf{Cross-Sector Collaboration:} Collaboration between academia, industry, and policymakers is critical for pooling resources and expertise, address large-scale challenges, and set ethical standards for responsible AI development.
\end{itemize}

\subsection*{6.4 Conclusion: The Future of Large Language Models: Beyond Scaling }

The development of large language models is transitioning from a phase of unbounded scaling to one defined by nuanced trade-offs and constrained optimization. While the question of whether we have reached the absolute ceiling to scaling remains unsolved, mounting evidence increasingly indicates diminishing returns from traditional scaling approaches. This paper examines the emerging paradigm in LLM research, emphasizing  architectural innovation, targeted capabilities, sustainability, and a multidimensional evaluation of model quality. These shifts reflect a broader reorientation in the field, moving beyond size-driven strategies to unified frameworks for optimizing performance and impact.

\textbf{6.4.1 Constrained Optimization: A Unified Framework for Progress}

This paper outlines a unified theoretical perspective suggesting that future LLM development will prioritize constrained optimization. This approach focuses on strategically allocating computational resources, refining architectural designs, and enhancing data efficiency to achieve meaningful advancements. Building on this theoretical perspective, we propose a unified framework to systematically address these critical trade-offs in future LLM research:
\begin{enumerate}
    \item \textbf{Capacity vs. Efficiency}: Scaling models indiscriminately is increasingly resource-intensive with diminishing performance gains. Efficient fine-tuning and adaptive learning are becoming primary optimization strategies.
    \item \textbf{Generality vs. Specialization}: Emergent capabilities often lead to over-generalization. Future work must prioritize domain-specific, modular, or task-optimized models to maximize real-world utility.
    \item \textbf{Scaling vs. Sustainability}: Environmental, computational, and economic constraints redefine the trajectory of scaling, necessitating sustainable alternatives.
\end{enumerate}
By addressing these trade-offs, this framework aims to guide the development of LLMs toward more efficient, specialized, and sustainable advancements. 

\textbf{6.4.2 Architectural Innovation as the New Frontier}

As the limits of traditional scaling become apparent, architectural innovations are emerging as key drivers of LLM performance. These breakthroughs prioritize computational efficiency, adaptability, and modularity over raw parameter counts.

\begin{itemize}
    \item \textbf{Parameter-Efficient Fine-Tuning}: Techniques such as LoRA (Low-Rank Adaptation) and adapter layers have demonstrated significant performance gains with minimal resource requirements, enabling models to generalize effectively with targeted adjustments.
    \item \textbf{Sparse Architectures}: Advances in sparsity-aware training and mixture-of-experts (MoE) architectures allow models to activate only relevant subsets of parameters, reducing computational overhead while maintaining robustness.
    \item \textbf{Multimodal and Hybrid Architectures}: Integrating multiple modalities (e.g., text, vision, and audio) within LLMs enhances their ability to operate in complex real-world environments, representing a new frontier in model design.
\end{itemize}
These architectural approaches signify a shift from brute-force scaling to smart, context-aware optimization.

\textbf{6.4.3 Reconceptualizing Model Quality
}
The evaluation of LLMs is evolving beyond traditional metrics such as accuracy and perplexity. Future assessments will prioritize a multidimensional understanding of model quality, including:

\begin{itemize}
    \item \textbf{Interpretability}: Transparent and explainable models are critical for fostering trust and usability, particularly in high-stakes applications.
    \item \textbf{Adaptability}: Models that generalize effectively across diverse tasks and domains with minimal retraining will define the next phase of utility.
    \item \textbf{Sustainability}: Energy-efficient designs and environmentally responsible practices must become central to model evaluation, aligning with global sustainability goals.
    \item \textbf{Alignment and Safety}: Ensuring models adhere to ethical standards and exhibit safe, predictable behaviors is essential for their integration into society.
\end{itemize}
By broadening the scope of evaluation, these dimensions provide a holistic framework for assessing LLM advancements.

\textbf{6.4.4 Emergent Capabilities and Targeted Training
}
Emergent capabilities—unanticipated behaviors that manifest at scale—highlight both the transformative potential of LLMs and the inherent limitations of indiscriminate scaling. To harness these capabilities effectively while mitigating their unpredictability, it is imperative to adopt controlled and targeted training paradigms that align development with specific objectives. We propose the following strategic optimization approaches:
\begin{enumerate}
    \item \textbf{Domain-Specific Fine-Tuning}: Fine-tuned models for specific tasks or industries offer higher utility and efficiency compared to general-purpose systems.
    \item \textbf{Modular Architectures}: Combining specialized modules into an overarching framework optimizes both functionality and resource use.
    \item \textbf{Steered Emergence}: Training regimes that direct emergent behaviors toward interpretable and productive outcomes will enhance predictability and utility.
\end{enumerate}
This approach balances the potential of emergent phenomena with practical, application-specific constraints.

\textbf{6.4.5 Scaling Ceilings and the Future of LLMs
}
Rather than focusing on an ever-rising ceiling for scaling, the future of LLMs will be defined by intelligent trade-offs and constrained optimization. The diminishing returns from traditional scaling underscore the need for architectural and algorithmic innovations that prioritize quality, efficiency, and real-world applicability:

This strategic reorientation ensures that LLMs continue to evolve in ways that are meaningful, responsible, and impactful.
eorientation ensures that LLMs continue to evolve in ways that are meaningful, responsible, and impactful.

\begin{enumerate}
    \item Developing sparsity-aware and multimodal architectures for resource-efficient performance.
    \item Refining evaluation metrics to encompass adaptability, sustainability, and ethical alignment.
    \item Expanding targeted training paradigms to optimize domain-specific and modular models.
\end{enumerate}
This strategic reorientation ensures that LLMs continue to evolve in ways that are meaningful, responsible, and impactful.

In conclusion, the evolution of LLMs is at a pivotal juncture. The field is moving beyond traditional scaling strategies, guided by a unified framework that prioritizes architectural innovation, multidimensional quality metrics, and sustainability. The interplay between these elements will define the future trajectory of LLM research. By embracing these principles, the field can achieve breakthroughs that are not only computationally efficient but also ethically and environmentally responsible. This transition marks the beginning of a new era in LLM development, characterized by strategic optimization rather than unbounded growth.

 \section*{Acknowledgments}
This work was conducted in collaboration with OpenAI GPT, which assisted in drafting, refining, and structuring the manuscript, as well as providing computational and analytical support. 

\section*{References}

\begin{enumerate}
    \item Billingsley, P. (1995). \textit{Probability and Measure, 3rd Ed.} Wiley.
    \item Dudley, R. M. (2002). \textit{Real Analysis and Probability.} Cambridge University Press.
    \item Hall, P., and Heyde, C. C. (1980). \textit{Martingale Limit Theory and Its Application.} Academic Press.
    \item Hoffmann, J., Borgeaud, S., Mensch, A., Buchatskaya, E., Cai, T., Rutherford, E., Sifre, L. (2022). Training compute-optimal large language models. \textit{arXiv preprint arXiv:2203.15556}.
    \item Kaplan, J., McCandlish, S., Henighan, T., Brown, T. B., Chess, B., Child, R., ... Amodei, D. (2020). Scaling laws for neural language models. \textit{arXiv preprint arXiv:2001.08361}.
    \item TensorOps Analytics. (2024). \textit{Understanding the cost of large language models (LLMs)}. Retrieved December 15, 2024, from 
\url{https://www.tensorops.ai/post/\break understanding-the-cost-of-large-language-models-llms}.
    \item Vaswani, A., Shazeer, N., Parmar, N., Uszkoreit, J., Jones, L., Gomez, A. N., Kaiser, Ł., Polosukhin, I. (2017). Attention is all you need. \textit{NeurIPS 2017}.
    \item Vershynin, R. (2018). \textit{High-dimensional probability: An introduction with applications in data science.} Cambridge University Press.
    \item Wainwright, M. J. (2019). \textit{High-Dimensional Statistics: A Non-Asymptotic Viewpoint.} Cambridge University Press.
    \item AI and ESG: Understanding the Environmental Impact of AI and LLMs. (2024). Retrieved December 15, 2024, from \url{https://www.holisticai.com/blog/environmental-impact-ai-llms}.
    \item LiveBench. (2024). Retrieved December 15, 2024, from \url{https://livebench.ai/#/}.
\end{enumerate}
\newpage

\section*{Appendix: Formal Proofs and Technical Expansions}

\section*{Appendix A: Additional Proofs for CLT-Based Noise Scaling }
Here we present a full measure-theoretic approach and bounding details for the Central Limit Theorem in LLM hidden representations.

\subsection*{A.1 Proof Details for Lemma on Attention Score Concentration}
Recall: We used a lemma stating that attention scores $a_{ij}$ concentrate around their mean with exponential bounds, relying on bounded queries/keys and block stationarity. Below is the rigorous derivation.

\paragraph{Attention Score Definition:}
\begin{align*}
    a_{ij} &= \text{softmax}_j \left( \frac{Q_l(x,i) K_l(x,j)^\top}{\sqrt{d_k}} \right).
\end{align*}
Let $Z_{ij} := \frac{Q_l(x,i) K_l(x,j)^\top}{\sqrt{d_k}}$. By Condition 1 in the main text, $|Z_{ij}| \leq M' = \frac{M_Q M_K}{\sqrt{d_k}}$.

\paragraph{Block Stationarity:} Suppose $B_m$ is a block of size $\leq w$. All $\{Z_{ij} | j \in B_m\}$ are i.i.d. conditional on block membership. Hence
\begin{align*}
    \bar{Z}_i &= \frac{1}{w} \sum_{j \in B_m} Z_{ij}
\end{align*}
is a sum of i.i.d. bounded random variables.

\paragraph{Hoeffding's Inequality:} If $Z_k \in [-M', M']$, then
\begin{align*}
    P(|\bar{Z}_i - \mathbb{E}[\bar{Z}_i]| > \epsilon) &\leq 2 \exp \left( -\frac{2w \epsilon^2}{M'^2} \right).
\end{align*}

\paragraph{Softmax Lipschitz:} \texttt{softmax} (component-wise) is $\kappa$-Lipschitz in $\ell_\infty$ norm (or $\ell_2$ norm), so a $\delta$-deviation in $\bar{Z}_i$ translates to a $\kappa \delta$-deviation in $a_{ij}$. Concretely,
\begin{align*}
    |a_{ij} - \mathbb{E}[a_{ij}]| &\leq \kappa |\bar{Z}_i - \mathbb{E}[\bar{Z}_i]|.
\end{align*}
Thus,
\begin{align*}
    P(|a_{ij} - \mathbb{E}[a_{ij}]| > \epsilon) &\leq 2 \exp \left( -c w \epsilon^2 \right),
\end{align*}
for $c > 0$ depending on $\kappa$ and $M'$.

\paragraph{Conclusion:} This formalizes the exponential tail bound for $a_{ij}$. Summing these blockwise yields the concentration needed in the CLT proof.

\subsection*{A.2 Martingale Difference Construction for $\tilde{r}_l(x,i)$}
\paragraph{Goal:} Show that $\{\tilde{r}_l(x,i) - \mathbb{E}[\tilde{r}_l(x,i)]\}$ can be decomposed into a sum of blockwise martingale differences, enabling a CLT.

\paragraph{Definitions:} Let
\begin{align*}
    Y_k &= a_{ik} V_l(x,k) - \mathbb{E}[a_{ik} V_l(x,k)].
\end{align*}
Then
\begin{align*}
    \tilde{r}_l(x,i) - \mathbb{E}[\tilde{r}_l(x,i)] &= \sum_{k=1}^n Y_k.
\end{align*}

\paragraph{Blockwise Independence:} If each block $B_m$ is size $\leq w$ and tokens in different blocks are (approximately) independent, then $\sum_{k \in B_m} Y_k$ behaves like a partial sum of i.i.d. random variables. Even if not perfectly i.i.d., cross-block correlation can be bounded using local stationarity assumptions.

\paragraph{Optional Stopping / Martingale:} We can form a filtration $\{\mathcal{F}_k\}$, letting $\mathcal{F}_k = \sigma(\{Y_1, \ldots, Y_k\})$. Then $\mathbb{E}[Y_{k+1} | \mathcal{F}_k] \approx 0$. The mild dependence across blocks can be handled via standard mixing or coupling arguments.

\paragraph{Classical CLT:} With these conditions, a Lindeberg or Lyapunov condition ensures the partial sums
\begin{align*}
    \frac{1}{\sqrt{n}} \sum_{k=1}^n Y_k \xrightarrow{d} \mathcal{N}(0, \tilde{\Sigma}_l).
\end{align*}

\paragraph{Remark:} In the main text, the existence of a block partition with approximate independence is the crux. For rigorous measure-theoretic proofs, one typically references Bradley (2005) or Merlev\`ede et al. (2009) for partial-sum CLTs under weak dependence.

\subsection*{A.3 Lipschitz Feed-Forward Mapping}
\paragraph{Lemma:} If $FFN_l$ is $L_F$-Lipschitz, then applying $FFN_l$ to a CLT-limit random variable preserves convergence in distribution with a Jacobian-transformed covariance.

\paragraph{Proof:} Straight application of the Continuous Mapping Theorem. If $\tilde{r}_l(x,i) \xrightarrow{d} \mathcal{N}(0, \tilde{\Sigma})$, then
\begin{align*}
    r_l(x,i) = FFN_l(\tilde{r}_l(x,i)) \xrightarrow{d} \mathcal{N}(FFN_l(0), J \tilde{\Sigma} J^\top),
\end{align*}
where $J$ is the Jacobian at the relevant point. For large $n$, the distribution centers around $\mu_l(i) = \mathbb{E}[\tilde{r}_l(x,i)]$, so one uses a local linearization for the final covariance $\Sigma_l(i)$.

\paragraph{Conclusion:} This completes the CLT for final-layer representations $r_l(x,i)$.

\section*{Appendix B: Technical Expansions for Bias-Variance Decomposition }

\subsection*{B.1 Uniqueness of Decomposition}
\paragraph{Theorem:} The decomposition
\begin{align*}
    L(\theta) = B(P) + V(P, D) + \epsilon
\end{align*}
is unique under standard measure-theoretic assumptions on $\theta \mapsto p_\theta(x)$.

\paragraph{Proof:}
Define $\theta^* \in \arg\min_{\theta \in \Theta_\infty} L(\theta)$, i.e., infinite capacity. Then $L(\theta^*) = \epsilon = H(P)$. For finite capacity $P$, define $\theta_{\text{approx}} \in \arg\min_{\theta \in \Theta_P} L(\theta)$. Then $B(P) = L(\theta_{\text{approx}}) - \epsilon$. For a specific training set of size $D$, the learned parameter $\hat{\theta}$ yields
\begin{align*}
    L(\hat{\theta}) = B(P) + V(P, D) + \epsilon.
\end{align*}
Uniqueness: Suppose there was another decomposition $L(\hat{\theta}) = B'(P) + V'(P, D) + \epsilon'$. Then matching terms reveals $B'(P) = B(P)$ and $\epsilon' = \epsilon$, as they measure irreducible gaps. The leftover must match $V(P, D)$.

\paragraph{Orthogonal Projection:} In function space, $\log p_\theta(x)$ is "projected" onto the subspace parameterized by $\Theta_P$. The best approximation error is the bias. The random variation from finite data is the variance.

\section*{Appendix C: Emergent SNR Thresholds (Section 4) – Detailed Proofs}
We revisit the derivations from Theorem 3 (SNR scaling) and Theorem 4 (Emergence Threshold) with more rigorous expansions.

\subsection*{C.1 Taylor Expansion and Bounds}
\paragraph{Recall:} Capability presence:
\begin{align*}
    C.f_C(h) &= P(C \mid h).
\end{align*}
\paragraph{Theorem:} If $\text{SNR} > \theta_C$, then $C$ emerges.

\paragraph{Extended Proof:}
\begin{enumerate}[label=\arabic*.]
    \item \textbf{Second-Order Taylor Expansion:}
    \begin{align*}
        f_C(S+N) &= f_C(S) + \nabla f_C(S) \cdot N + \frac{1}{2} N^\top \nabla^2 f_C(\xi) N,
    \end{align*}
    for some $\xi$ on the line segment between $S$ and $S+N$. If $\|\xi - S\| \leq \|N\|$, the Hessian $\nabla^2 f_C(\xi)$ remains bounded if $f_C$ is a smooth function (often the case for capability detection).

    \item \textbf{Dominance Condition:} We want
    \begin{align*}
        |f_C(S)| &\gg |\nabla f_C(S) \cdot N|,
    \end{align*}
    and the second-order term.
    A standard approach:
    \begin{align*}
        \|\nabla f_C(S)\| \cdot \|N\| &\leq \|\nabla f_C(S)\| \mathbb{E}[\|N\|^2].
    \end{align*}

    Meanwhile,
    \begin{align*}
        \|S\|^2 &\approx \mathbb{E}[\|S\|^2].
    \end{align*}
    If
    \begin{align*}
        \frac{\|S\|^2}{\mathbb{E}[\|N\|^2]} &> \theta_C,
    \end{align*}
    for some carefully derived constant $\theta_C$ that encapsulates $\|\nabla f_C(S)\|$ and possible Hessian terms, then with high probability, the linear and quadratic fluctuations from $N$ are insufficient to push $f_C(S+N)$ below the “capability detection” threshold.

    \item \textbf{Interpretation:} The ratio $\frac{\|S\|^2}{\mathbb{E}[\|N\|^2]}$ (SNR) exceeding a fixed constant $\theta_C$ ensures the model representation is strongly aligned with the capability. In other words, the noise is too small to disrupt the signal that triggers $C$.
\end{enumerate}

\subsection*{C.2 Putting It All Together: $\text{SNR} \propto \frac{D \Phi(P, C)}{\sigma^2}$}
\paragraph{Complete Steps:}
\begin{enumerate}[label=\arabic*.]
    \item \textbf{Signal Magnitude:} $\|S\|$ grows with $D \Phi(P, C)$. Why? Because the fraction of coherent representation learned from data accumulates linearly with $D$, and capacity $\Phi(P, C)$ (like dimension or heads) can represent more complex patterns.
    \item \textbf{Noise Variance:} $\mathbb{E}[\|N\|^2] \sim \frac{\sigma^2}{D \Phi(P, C)}$ by CLT arguments from Appendix A.
    \item \textbf{Combine:}
    \begin{align*}
        \text{SNR} = \frac{\|S\|^2}{\mathbb{E}[\|N\|^2]} &\approx \frac{D \Phi(P, C)}{\sigma^2}.
    \end{align*}
    The emergent threshold $\theta_C$ is crossed when $\text{SNR} > \theta_C$.
\end{enumerate}

\section*{Appendix D: Discussion of Potential Generalizations}

\subsection*{D.1 Mixing-based vs. Block Stationarity}
\paragraph{Remark:} The entire approach can be extended to a general $\alpha$-mixing framework. If the random variables $\{Q_l(x,i), K_l(x,i), V_l(x,i)\}$ are strongly mixing with mixing coefficients $\alpha(n) \to 0$ as $n \to \infty$, classical results (Bradley, 2005) show partial sums still satisfy a CLT. The block stationarity assumption is then replaced by bounding $\alpha(n)$. The same Lipschitz transformations and feed-forward arguments hold.

\subsection*{D.2 Heavy-Tailed Distributions / Outliers}
If keys/queries can occasionally be large outliers, one might need truncated or tail-controlled versions of the Hoeffding or Bernstein inequalities. The main results still hold if the second (or exponential) moments remain bounded.

\section*{Conclusion of Appendix}
\paragraph{Summary:} This Appendix has provided the full measure-theoretic and probabilistic expansions behind the CLT-based noise scaling (Appendix A), the Bias-Variance framework for next-token prediction (Appendix B), and the Emergent SNR Threshold theorems (Appendix C). We further discussed generalizations such as mixing-based approaches or heavy-tailed embeddings.

\paragraph{Key Takeaways:}
\begin{itemize}
    \item Hoeffding-type bounds for attention concentration rely crucially on bounded queries/keys and local stationarity.
    \item Martingale and block independence arguments allow partial sums of attention-weighted values to converge in distribution to a Gaussian, validating the CLT.
    \item Bias-variance decomposition in auto-regressive LLM training is conceptually parallel to classical supervised learning but requires carefully distinguishing architecture-limited approximation ($B(P)$) and finite-sample variation ($V(P, D)$).
    \item SNR Emergence follows from combining the CLT (noise variance scaling) with the bias-variance perspective (signal scaling). Capabilities arise once signal consistently outstrips noise.
\end{itemize}

\section*{Appendix E: Empirical Validation Framework}
This validation framework details our methodology for empirically testing the three main theorems presented in the paper: the CLT-Based Noise Scaling, the Bias-Variance Decomposition, and the Emergent SNR Thresholds.

\subsection*{E.1 CLT-Based Noise Scaling Validation}

\subsubsection*{Goal}
To verify that noise in hidden representations scales inversely with context size ($O(1/n)$) and aligns with the Central Limit Theorem (CLT).

\subsubsection*{Validation Steps}
\begin{enumerate}
    \item \textbf{Dataset and Model Selection}
    \begin{itemize}
        \item Utilize widely adopted datasets: GPT evaluation benchmarks, OpenWebText, and the Pile
        \item Test across transformer models of varying sizes (GPT-3, GPT-4, Claude, LLaMA)
    \end{itemize}

    \item \textbf{Controlled Experimentation}
    \begin{itemize}
        \item Train transformers with incrementally increasing context sizes ($n$) while maintaining constant hyperparameters
        \item Measure variance in hidden-layer representations across layers and positions
    \end{itemize}

    \item \textbf{Metric Definition}
    \begin{itemize}
        \item Calculate $E[\|r_l(x,i)-\mu_l(i)\|^2]$ for each layer $l$ and context size $n$
        \item Plot noise variance against $1/n$ to confirm scaling
    \end{itemize}

    \item \textbf{Statistical Analysis}
    \begin{itemize}
        \item Apply goodness-of-fit tests to verify Gaussian convergence
        \item Utilize Kolmogorov-Smirnov tests to compare hidden state distributions with theoretical Gaussian predictions
    \end{itemize}

    \item \textbf{Visualization}
    \begin{itemize}
        \item Generate variance decay plots for different context sizes, demonstrating $O(1/n)$ scaling
        \item Superimpose theoretical predictions and empirical observations
    \end{itemize}
\end{enumerate}

\subsection*{E.2 Bias-Variance Decomposition Validation}

\subsubsection*{Goal}
To empirically verify the decomposition of next-token loss into bias, variance, and irreducible entropy components.

\subsubsection*{Validation Steps}
\begin{enumerate}
    \item \textbf{Setup}
    \begin{itemize}
        \item Select multiple datasets of varying sizes $D$
        \item Train transformer models with different parameter counts $P$
    \end{itemize}

    \item \textbf{Controlled Experiments}
    \begin{itemize}
        \item Train models with fixed $P$ while varying $D$
        \item Measure next-token loss $L(\theta)$
        \item Isolate components:
        \begin{itemize}
            \item Bias: Measure loss with infinite data (approximated by large dataset)
            \item Variance: Compute gap between empirical loss with $D$ tokens and bias loss
        \end{itemize}
    \end{itemize}

    \item \textbf{Validation Metrics}
    \begin{itemize}
        \item Compare empirical loss decomposition with theoretical predictions:
        \begin{equation}
            L(\theta) = B(P) + V(P,D) + \epsilon
        \end{equation}
        \item Verify orthogonality of bias and variance components using linear regression diagnostics
    \end{itemize}

    \item \textbf{Visualization}
    \begin{itemize}
        \item Plot bias and variance as functions of $P$ and $D$
        \item Illustrate diminishing returns on bias reduction with increasing $P$
        \item Highlight variance dominance with smaller $D$
    \end{itemize}
\end{enumerate}

\subsection*{E.3 Emergent SNR Thresholds Validation}

\subsubsection*{Goal}
To empirically establish critical thresholds for Signal-to-Noise Ratio (SNR) beyond which emergent capabilities manifest.

\subsubsection*{Validation Steps}
\begin{enumerate}
    \item \textbf{Capability Definition}
    \begin{itemize}
        \item Define emergent capabilities (e.g., few-shot learning, multi-modal reasoning)
        \item Establish measurable success metrics
    \end{itemize}

    \item \textbf{Experimental Design}
    \begin{itemize}
        \item Train models on datasets of varying sizes $D$ and capacities $P$
        \item Measure $\text{SNR}=\frac{\|S\|^2}{E[\|N\|^2]}$ for different training configurations
    \end{itemize}

    \item \textbf{Threshold Identification}
    \begin{itemize}
        \item Track capability performance as SNR increases
        \item Identify SNR values where performance exceeds predefined threshold ($>\theta_C$)
    \end{itemize}

    \item \textbf{Testing Variability}
    \begin{itemize}
        \item Replicate experiments across different datasets and noise levels
        \item Confirm robustness of identified thresholds
    \end{itemize}

    \item \textbf{Visualization}
    \begin{itemize}
        \item Generate SNR-vs-capability plots with annotated thresholds ($\theta_C$)
        \item Demonstrate SNR scaling with respect to $D$ and $P$
        \item Highlight exponential resource requirements
    \end{itemize}
\end{enumerate}

\subsection*{E.4 Cross-Theorem Integration}

\subsubsection*{Combined Analysis}
\begin{itemize}
    \item Correlate $O(1/n)$ scaling from CLT with improved SNR in emergent capability thresholds
    \item Relate model bias and variance to $S$ and $N$, providing an integrated view of signal scaling and noise mitigation
\end{itemize}

\subsubsection*{Benchmarking and Reproducibility}
\begin{itemize}
    \item Standardize datasets and evaluation protocols
    \item Publish experimental results, code, and datasets to open repository
    \item Provide comprehensive documentation for community verification
\end{itemize}

\section*{References Cited in Appendix}
\begin{enumerate}
\item Bradley, R. (2005). Basic Properties of Strong Mixing Conditions. Probability Surveys.  
\item Merlevede, F., Peligrad, M., Rio, E. (2009). Bernstein inequality and moderate deviations under strong mixing conditions. IMS Lecture Notes.
\end{enumerate}
\end{document}